\documentclass{article}
\usepackage{ijcai24}

\usepackage{times}
\usepackage{soul}
\usepackage{url}
\usepackage[hidelinks]{hyperref}
\usepackage[utf8]{inputenc}
\usepackage[small]{caption}
\usepackage{graphicx}
\usepackage{amsmath}
\usepackage{amsthm}
\usepackage{algorithm}
\usepackage{algorithmic}
\usepackage[switch]{lineno}
\usepackage{times}
\usepackage{latexsym}
\usepackage{graphicx}
\usepackage{amsmath}
\usepackage{tabularx} 
\usepackage{multirow} 
\usepackage{leftidx}
\usepackage{hyperref}

\usepackage{enumitem}
\usepackage{tabularx}
\usepackage{xcolor}         
\usepackage{graphicx}
\usepackage{amsmath} 
\usepackage{amsfonts}       
\usepackage{nicefrac}       
\usepackage{microtype}      
\usepackage{booktabs}       

\title{Sentence-Level or Token-Level? A Comprehensive Study on Knowledge Distillation}

\author{
Jingxuan Wei$^{1,2\ast}$ \and
Linzhuang Sun$^{1,2}$ \and
Yichong Leng$^3$ \and
Xu Tan$^{4 \ast}$ \and
Bihui Yu$^{1,2}$ \and
Ruifeng Guo$^{1,2}$
\\
\affiliations
$^1$Shenyang Institute of Computing Technology, Chinese Academy of Sciences\\
$^2$University of Chinese Academy of Sciences\\
$^3$University of Science and Technology of China\\
$^4$Independent Researcher\\
\emails
weijingxuan20@mails.ucas.edu.cn, tanxu2012@gmail.com
}

\begin{document}

\maketitle

\begin{abstract}

Knowledge distillation, transferring knowledge from a teacher model to a student model, has emerged as a powerful technique in neural machine translation for compressing models or simplifying training targets.
Knowledge distillation encompasses two primary methods: sentence-level distillation and token-level distillation.
In sentence-level distillation, the student model is trained to align with the output of the teacher model, which can alleviate the training difficulty and give student model a comprehensive understanding of global structure. Differently, token-level distillation requires the student model to learn the output distribution of the teacher model, facilitating a more fine-grained transfer of knowledge.
Studies have revealed divergent performances between sentence-level and token-level distillation across different scenarios, leading to the confusion on the empirical selection of knowledge distillation methods. In this study, we argue that token-level distillation, with its more complex objective (i.e., distribution), is better suited for ``simple'' scenarios, while sentence-level distillation excels in ``complex'' scenarios. To substantiate our hypothesis, we systematically analyze the performance of distillation methods by varying the model size of student models, the complexity of text, and the difficulty of decoding procedure.
While our experimental results validate our hypothesis, defining the complexity level of a given scenario remains a challenging task. So we further introduce a novel hybrid method that combines token-level and sentence-level distillation through a gating mechanism, aiming to leverage the advantages of both individual methods. Experiments demonstrate that the hybrid method surpasses the performance of token-level or sentence-level distillation methods and the previous works by a margin, demonstrating the effectiveness of the proposed hybrid method.
\end{abstract}

\section{Introduction}
\label{introduction}
Knowledge distillation, as a fundamental technique for model compression and knowledge transfer in deep neural networks, has wide application in the field of neural machine translation (NMT) ~\cite{hinton2015distilling,gou2021knowledge}. Knowledge distillation involves transferring knowledge from a larger, cumbersome model to a smaller, more efficient one, serving purposes such as compressing machine translation models and simplifying training targets for non-autoregressive models ~\cite{phuong2019towards,liu2020adaptive,wang2021knowledge,xiao2023survey}.

Given the variance in training targets, knowledge distillation in NMT can be divided into two main categories: sentence-level knowledge distillation and token-level knowledge distillation. Sentence-level knowledge distillation mainly focuses on simplifying the training target to improve the translation accuracy ~\cite{DBLP:conf/acl/GajbhiyeFABOAS21,DBLP:conf/naacl/YangS022}. Specifically, given a source and target sentence pair, sentence-level distillation firstly feeds the source sentence into the teacher model to generate a pseudo target sentence, then the pseudo target sentence is leveraged as the training target of student model. Compared with the origin target sentence, the distribution of pseudo target sentence is simpler, and thus easier to learn for student model~\cite{kim2016sequence,zhang2019future,tang2019distilling,tan2022document}. %

In contrast, token-level knowledge distillation focuses on enhancing translation quality by a finer granularity ~\cite{kim2016sequence,DBLP:conf/icassp/MunimIS19}. Different with sentence-level knowledge distillation which only leverages the output sentence of teacher model, token-level knowledge distillation further uses the token distribution in the output sentence. The student model is trained to output a similar distribution with the teacher model on every token, which helps the student model learn detail knowledge on token difference and be more suitable for texts with high lexical diversity~\cite{DBLP:conf/acl/WangJBWHT20}.

However, empirical studies have revealed divergent performances between sentence-level and token-level distillation across different scenarios. Specifically, while some scenarios benefit more from the global structure and semantic consistency provided by sentence-level distillation~\cite{kim2016sequence,chen2020distilling,xu2021does,lei2022sentence,m2023multilingual}, other scenarios require the fine-grained knowledge transfer that token-level distillation offers~\cite{liao2020multi,tang2021vidlankd,li2021metats,ma2023multi}. This variation in performance has led to confusion regarding the empirical selection of knowledge distillation methods. In this study, we conduct analytical experiments to explore the general suitable scenario of two knowledge distillation methods. Given that the training target of sentence-level distillation (simplified sentence by teacher model) is easier than that of the token-level distillation (detailed token distribution of teacher model). We hypothesize that sentence-level distillation is suitable for ``complex'' scenarios and the token-level distillation is suitable for ``simple'' scenarios.

We define the ``complex'' or ``simple'' scenarios from three perspectives: 1) model size of student model, as the student model becomes small, it is harder for the student model to learn the knowledge, and thus the  scenario become more complex; 2) the complexity of text, more complex text will make the learning procedure of student model harder; 
3) the difficulty of decoding, which is determined by the amount of auxiliary information available during decoding. The more auxiliary information available, the simpler the decoding process becomes.
Experiments on the above three perspectives consistently verify our hypothesis, showing that the token-level distillation performs better in simple scenarios with the sentence-level distillation is better for complex scenarios. 

Although the analytical experiments provide deep understanding and reveal the general suitable scenarios for two distillation method, how to empirically define the complexity of a machine translation task is challenging. 
To address this challenge, we further explore the hybridization of two distillation methods, aiming at taking the advantage of both the distillation methods to enhance overall translation accuracy. We propose a dynamic gating mechanism that adaptively balances the learning process between sentence-level and token-level distillation. Specifically, the student model is trained to learn both the pseudo target sentence distribution for global coherence and the detailed token distribution from the teacher model for lexical precision, with the gating mechanism dynamically adjusting the emphasis based on the evolving learning context and model performance.

The contributions of this paper are summarized as follows:
\begin{itemize}
\item 
We conduct experiments to discover the optimal use of sentence-level and token-level distillation, uncovering that the sentence-level distillation excels in simpler scenarios, whereas the token-level distillation is more effective in complex ones.
\item 
We propose a hybrid method of sentence-level and token-level distillation, showing enhanced performance over single distillation methods and baseline models.
\end{itemize}

\section{Related Work}
\label{gen_inst}
Knowledge distillation (KD) is widely applied in the field of neural machine translation (NMT) to enhance the efficiency and performance of translation models~\cite{hinton2015distilling,xu2021does,chen2020distilling,gou2021knowledge,zhang2022fine}.
Recently, knowledge distillation is applied in multilingual NMT~\cite{DBLP:conf/iclr/TanRHQZL19} to assist models in mastering multiple languages within a single framework.
A multi-agent learning framework~\cite{liao2020multi} is utilized to investigate how sentence-level and token-level distillation can work together synergistically.
PMGT~\cite{ding2021progressive}enhances phrase translation accuracy and model reordering capability by progressively increasing the granularity of training data from words to sentences.
ProKD~\cite{ge2023prokd} demonstrates the use of high-resource language teacher models to enhance translation performance in low-resource languages through cross-lingual knowledge distillation.
Despite the emergence of various distillation models, knowledge distillation in NMT, from the perspective of training targets, can be primarily divided into two categories: sentence-level knowledge distillation and token-level knowledge distillation.

\subsection{Token-Level Knowledge Distillation}
Token-level knowledge distillation in neural machine translation (NMT) primarily focuses on enhancing the translation accuracy of individual words or phrases~\cite{he2021distiller,gou2021knowledge,wang2021knowledge}. This approach is explored in various studies to improve specific aspects of translation quality. For example, token-level ensemble distillation for grapheme-to-phoneme conversion~\cite{DBLP:conf/interspeech/SunTGLZQL19} can enhance the phonetic translation accuracy. Additionally, a selective knowledge distillation method~\cite{wang2021selective} aims at optimizing the word-level distillation loss and the standard prediction loss. The raw data exposure model~\cite{ding2020understanding} reduces lexical choice errors in low-frequency words by exposing NAT models to raw data, enhancing translation accuracy. SKD~\cite{sun2020knowledge} investigates knowledge distillation in the context of multilingual unsupervised NMT, while kNN-KD~\cite{yang2022nearest} examines the effects of nearest-neighbor knowledge distillation on translation accuracy. Furthermore, the token-level self-evolution training~\cite{peng2023token} method dynamically identifies and focuses on under-explored tokens to improve lexical accuracy, generation diversity, and model generalization. The concept of knowledge distillation via token-level relationship graphs~\cite{zhang2023knowledge} offers a novel perspective on leveraging relational data for distillation, further contributing to the advancement of the token-level knowledge distillation in NMT.

\subsection{Sentence-Level Knowledge Distillation}
Sentence-level knowledge distillation in neural machine translation (NMT) focuses on reducing the training difficulty of student model, particularly useful in capturing the semantics of whole sentences or long sequences~\cite{kim2016sequence,ren2019fastspeech,stahlberg2020neural}. For examples, ensemble distillation method~\cite{freitag2017ensemble} is proposed to effectively combine multiple model outputs to improve the handling of complex sentence structures. The scope of sentence-level distillation techniques is further expanded with the help of perturbed length-aware position encoding in non-autoregressive neural machine translation~\cite{oka2021using}. 
DDRS~\cite{shao2022one} introduces diversified distillation and reference selection strategies to improve the accuracy of sentence-level distillation.
Sentence-level distillation is also employed for simultaneous machine translation to address the challenges of real-time translation~\cite{deng2023improving}.

\begin{table*}[t]
\centering
\caption{Impact of model size on knowledge distillation across datasets. The $\bigtriangleup$ column represents the difference between token-level and sentence-level BLEU scores. Positive values suggest that the token-level distillation has a higher BLEU score than the sentence-level distillation. }
\label{tab:combined_impact_of_model_capacity}
\begin{tabular}{ccccccc}
\toprule
\multirow{2}{*}{Dataset} & \multirow{2}{*}{Teacher Size} & \multirow{2}{*}{Student Size} & \multicolumn{4}{c}{BLEU Score} \\
\cmidrule{4-7}
& & & Teacher Results & Token-level & Sentence-level &  $\bigtriangleup$  \\
\midrule
\multirow{4}{*}{IWSLT14 de$\rightarrow$en} & \multirow{4}{*}{38M} & 3M &  \multirow{4}{*}{34.80}  & 30.50 & 31.09 & -0.59 \\
& & 9M && 34.12 & 34.20 & -0.08 \\
& & 38M && 36.09 & 34.84 & 1.25 \\
& & 111M &  & 36.40 & 34.87 & 1.53 \\
\midrule
\multirow{4}{*}{IWSLT13 en$\rightarrow$fr} & \multirow{4}{*}{52M} & 7M & \multirow{4}{*}{44.10} & 39.63 & 41.94 & -2.31 \\
& & 12M &  & 42.42 & 43.48 & -1.06 \\
& & 52M &  & 44.82 & 44.43 & 0.39 \\
& & 140M &  & 44.87 & 44.26 & 0.61 \\
\midrule
\multirow{4}{*}{WMT14 en$\rightarrow$de} & \multirow{4}{*}{83M} & 28M &  \multirow{4}{*}{27.35} & 23.89 & 25.17 & -1.28 \\
& & 83M & & 26.49 & 26.77 & -0.28 \\
& & 112M &  & 26.73 & 26.68 & 0.05 \\
& & 146M &  & 26.66 & 26.56 & 0.10 \\
\midrule
\multirow{4}{*}{IWSLT17 ar$\rightarrow$en} & \multirow{4}{*}{47M} & 13M &  \multirow{4}{*}{31.19} & 28.66 & 30.21 & -1.55 \\
& & 24M  &  & 29.02 & 30.52  & -1.50 \\
& & 47M &  & 32.18 & 31.15  & 1.03 \\
& &84M &  & 32.37 & 31.33 & 1.04 \\
\bottomrule
\end{tabular}
\end{table*}

Several studies have provided insights to better understand the knowledge distillation. For instance, NAT~\cite{DBLP:conf/iclr/ZhouGN20} delves into why knowledge distillation is effective in non-autoregressive machine translation (NAT), uncovering the impact of text complexity on NAT. However, this study does not explore how text complexity affects token-level and sentence-level distillation.
HKD~\cite{lee2022hard} investigates the question of ``when to distill such knowledge''. It proposes a gate knowledge distillation scheme, where the teacher model serves not only as a knowledge provider but also as a calibration measurement, allowing for a switch between learning from the teacher model and training the student. This work also investigates both token-level and sentence-level distillation in teacher model. However, it treats them as separate strategies with independent token-level and sentence-level gates and fails to combine these two approaches.
Our work explores the general suitable scenario of knowledge distillation for both token-level and sentence-level perspectives, hypothesizing that token-level distillation is better suited for `simple' scenarios, while sentence-level distillation excels in `complex' scenarios. Furthermore, we propose a hybrid method that combines token-level and sentence-level distillation through a gating mechanism, aiming to alleviate the empirical confusion on selecting the distillation methods.

\section{Comprehensive Analysis of Knowledge Distillation}
\label{sec:analysis}
This section presents a detailed analysis of knowledge distillation within neural machine translation (NMT), focusing on the empirical evaluation of token-level versus sentence-level distillation in varied scenarios. This analysis aligns with our hypothesis outlined in Section \ref{introduction}: that sentence-level distillation is more adept in 'complex' scenarios, while token-level distillation excels in 'simple' scenarios.
We define the complexity from three perspectives: 

1) Model size of the student model: The scenarios become more complex when the model size of student model become smaller, since the student model need to compress the knowledge of teacher model into a model with limited capacity.

2) Complexity of the text: Datasets with more complex text, characterized by intricate sentence structures and diverse vocabulary, present more challenging learning environments for the student model.

3) Difficulty of decoding: The decoding difficulty is determined by the amount of ground truth or auxiliary information available during decoding. Scenarios where the decoder receives more ground truth or auxiliary information are considered simpler, as this additional information not only simplifies the decoding process by providing clearer guidance and reducing ambiguity, but also helps in avoiding the accumulation of errors during the decoding procedure.

In the following subsection, we firstly introduce the dataset and configuration used in the analysis experiments, then we verify our hypothesis from the above three perspectives.

\subsection{Dataset and Configuration}
For the experiments, we select four datasets to cover a range of complexities and linguistic characteristics: IWSLT13 English$\rightarrow$French (en$\rightarrow$fr), IWSLT14 German$\rightarrow$English (de$\rightarrow$en), WMT14 English$\rightarrow$German (en$\rightarrow$de), and IWSLT17 Arabic$\rightarrow$English (ar$\rightarrow$en). Each dataset offers a unique combination of bilingual sentence pairs and complexity levels: 200k for IWSLT13 en$\rightarrow$fr, 153k for IWSLT14 de$\rightarrow$en, 4.5M for WMT14 en$\rightarrow$de, and 231k for IWSLT17 ar$\rightarrow$en. 

\begin{table*}[t]
\centering
\caption{Impact of text complexity on knowledge distillation across datasets. The $\bigtriangleup$ column represents the difference between token-level and sentence-level BLEU scores. The $\bigtriangleup$ Rate (T) and $\bigtriangleup$ Rate (S) columns represent the percentage decrease in BLEU scores from the original to moderate and high noise levels for token-level and sentence-level respectively.}
\label{tab:combined_impact_of_data_complexity_extended}
\begin{tabular}{cccccccc}
\toprule
\multirow{2}{*}{Dataset} & \multirow{2}{*}{Stud Size} & \multirow{2}{*}{Noise} & \multicolumn{5}{c}{BLEU Score} \\
\cmidrule{4-8}
& & & Token & Sentence & $\bigtriangleup$ & $\bigtriangleup$ Rate (T) & $\bigtriangleup$ Rate (S) \\
\midrule
\multirow{3}{*}{IWSLT14 de$\rightarrow$en} & \multirow{3}{*}{38M} & Orig & 36.09 & 34.84 & 1.25 & - & - \\
& & Mod & 34.31 & 33.68 & 0.63 & -4.93\% & -3.33\% \\
& & High & 32.71 & 33.26 & -0.55 & -9.37\% & -4.54\% \\
\midrule
\multirow{3}{*}{IWSLT13 en$\rightarrow$fr} & \multirow{3}{*}{18M} & Orig & 44.56 & 43.95 & 0.61 & - & - \\
& & Mod & 42.89 & 42.50 & 0.39 &-3.75\% & -3.30\% \\
& & High & 41.11 & 42.53 & -1.42 & -7.74\% & -3.23\% \\
\midrule
\multirow{3}{*}{WMT14 en$\rightarrow$de} & \multirow{3}{*}{112M} & Orig & 26.73 & 26.68 & 0.05 & - & - \\
& & Mod & 25.03 & 25.47 & -0.44 & -6.36\% & -4.54\% \\
& & High & 24.49 & 25.35 & -0.86 & -8.38\% & -4.99\% \\
\midrule
\multirow{3}{*}{IWSLT17 ar$\rightarrow$en} & \multirow{3}{*}{47M} & Orig & 32.18 & 31.15 & 1.03 & - & - \\
& & Mod & 30.24 & 30.15 &  0.09 & -6.03\% &  -3.21\% \\
& & High & 27.90 & 28.23  &  -0.33 & -13.30\%  & -9.37\%  \\
\bottomrule
\end{tabular}
\end{table*}

We apply byte-pair encoding (BPE) with subword-nmt toolkit\footnote{https://github.com/rsennrich/subword-nmt} to all sentences in these datasets for tokenization. The vocabulary size is 32K.
The experiments are conducted using the Fairseq\footnote{\url{https://github.com/facebookresearch/fairseq}} framework.

\subsection{Impact of Model Size}
\label{subsec:model_capacity}
In this subsection, we explore the impact of student model size on the effectiveness of token-level and sentence-level distillation. 
We adjust the size of the student model following the model size reduction approach in~\cite{DBLP:conf/iclr/ZhouGN20} to observe the impact of model size on knowledge distillation across different datasets. The results are shown in Table~\ref{tab:combined_impact_of_model_capacity}.

\subsubsection{Analysis of Results and Summary}
\label{subsec:capacity_analysis_summary}

Our comprehensive analysis, as detailed in Table \ref{tab:combined_impact_of_model_capacity}, reveals a clear relationship between the student model's size and the effectiveness of knowledge distillation methods. Across all datasets, we observed a consistent trend: as the model size increases, both token-level and sentence-level distillation methods show improvement in BLEU scores. This improvement is particularly notable in the transition from small to medium-sized models. For instance, in the IWSLT14 de$\rightarrow$en dataset, a significant leap in performance is observed when the model size was increased from 3M to 9M parameters. However, beyond a certain threshold, such as 38M parameters in this dataset, the rate of improvement begins to plateau, indicating diminishing returns with further increases in size.

Interestingly, a critical point of inversion is observed where the advantage shifts from sentence-level to token-level distillation as the model size increases. In smaller models, sentence-level distillation tends to outperform token-level, aligning with our hypothesis that it is more suitable for complex scenarios where model size is limited. As the size increases, token-level distillation begins to show a relative advantage, suggesting its effectiveness in simpler scenarios with larger model capacities.

This trend suggests that while larger models can benefit from both distillation methods, there is an optimal range of model size where the gains are most substantial. Beyond this range, the additional complexity of larger models does not translate into proportional improvements in distillation performance. In practical terms, this implies that for scenarios prioritizing model compression, such as deploying NMT systems on resource-constrained devices, sentence-level distillation is more suitable due to its effectiveness in smaller models. Conversely, in scenarios where the focus is on maximizing translation accuracy, such as in server-based applications with fewer computational constraints or competition scenario~\cite{farinha2022findings,blain2023findings}, token-level distillation becomes increasingly advantageous as model size grows.

\begin{table*}[t]
\centering
\caption{Impact of decoding difficulty on BLEU scores: comparing Beam Search (BS) and Teacher Forcing (TF) methods. `BS-Token' and `BS-Sentence' represent BLEU scores using beam search for token-level and sentence-level distillation, respectively. `TF-Token' and `TF-Sentence' denote BLEU scores using teacher forcing for token-level and sentence-level distillation. $\bigtriangleup$BS and $\bigtriangleup$TF represent the differences in BLEU scores between token-level and sentence-level distillation for beam search and teacher forcing methods, respectively.}
\label{tab:combined_impact_of_Decoding_Difficulty}
\begin{tabular}{cccccccc}
\toprule
\multirow{2}{*}{Dataset} & \multirow{2}{*}{Stud Size} & \multicolumn{6}{c}{BLEU Score} \\
\cmidrule{3-8}
& & BS-Token & BS-Sentence & $\bigtriangleup$BS & TF-Token & TF-Sentence & $\bigtriangleup$TF \\
\midrule
\multirow{1}{*}{IWSLT14 de$\rightarrow$en} & 3M & 30.50 & 31.09 & 0.59 & 34.16 & 33.50 & -0.66 \\
\midrule
\multirow{1}{*}{IWSLT13 en$\rightarrow$fr} & 12M & 42.42 & 43.48 & 1.06 & 45.97 & 45.29 & -0.68 \\
\midrule
\multirow{1}{*}{WMT14 en$\rightarrow$de} & 83M & 26.49 & 26.77 & 0.28 & 29.82 & 28.58 & -1.24 \\
\midrule
\multirow{1}{*}{IWSLT17 ar$\rightarrow$en} & 47M & 32.18 & 31.15 & 1.03 & 32.51 & 31.60  & -0.91 \\
\bottomrule
\end{tabular}
\end{table*}

\subsection{Impact of Text Complexity}
\label{subsec:data_complexity}
In this subsection, we investigate the impact of text complexity, reflected by the presence of noise, on token-level and sentence-level distillation. Using IWSLT14 de$\rightarrow$en, IWSLT13 en$\rightarrow$fr, WMT14 en$\rightarrow$de, and IWSLT17 ar$\rightarrow$en datasets, we aim to understand how various levels of noise influence the effectiveness of each distillation approach.

\subsubsection{Experimental Setup and Methodology}
\label{subsec:exp_setup_methodology}
To assess the impact of text complexity on knowledge distillation, we introduce varying levels of noise to the datasets. We follow the methodology in~\cite{DBLP:conf/emnlp/EdunovOAG18}, applying three conditions to each dataset: \textit{no} noise, \textit{moderate} noise, and \textit{high} noise. We introduce the noise through token manipulation including deletion, substitution, and swapping.

Specifically, under \textit{moderate} noise conditions, we randomly delete and substitute 10\% of the tokens and conduct token swapping with a 50\% probability, maintaining a swap length of 3. This setup aims to simulate real-world linguistic processing errors and syntactic disarray.
For \textit{high} noise conditions, we keep the token deletion and substitution probabilities unchanged but increase the token swapping probability to 100\%, further elevating syntactic complexity. 
Our implementation of these manipulations references the methods available in this resource\footnote{\url{https://github.com/valentinmace/noisy-text/tree/e73c83dd1f08c25210c27abebf74d304de0d24e2}}.
In our experiments, the teacher models are Transformer-based, consistent with those in Table \ref{tab:combined_impact_of_model_capacity}, using the default sizes in Fairseq~\cite{ott2019fairseq} for each dataset. Our analysis focuses on comparing results under different noise conditions to evaluate the impact of text complexity on distillation effectiveness. The results are shown in Table~\ref{tab:combined_impact_of_data_complexity_extended}.

\subsubsection{Analysis of Results and Summary}
\label{subsec:results_summary}

From the results in Table \ref{tab:combined_impact_of_data_complexity_extended}, we observe a trend across all datasets: as the text complexity increases, both token-level and sentence-level distillation show a decrease in performance. However, sentence-level distillation demonstrates greater resilience, evidenced by a generally smaller decline in BLEU scores compared to token-level distillation, particularly in high noise scenarios. This is reflected from the lower average $\bigtriangleup$ Rate (S) across different noise levels, indicating its suitability for handling complex text scenarios.
In contrast, token-level distillation exhibits a more significant performance drop with the increased text complexity, as shown by the higher $\bigtriangleup$ Rate (T). 

In general, when the noise is low, the token-level distillation shows higher accuracy than the sentence-level distillation (negative $\bigtriangleup$ values in \textit{Orig} noise setting in Table~ \ref{tab:combined_impact_of_data_complexity_extended}). As the noise become higher, student models trained with sentence-level distillation display a better performance than those with token-level distillation (positive $\bigtriangleup$ values in \textit{High} noise setting in Table~ \ref{tab:combined_impact_of_data_complexity_extended}). The above phenomenon aligns with our hypothesis that token-level distillation is more effective in simpler scenarios with lower text complexity.

These findings highlight the importance of text complexity in the selection of appropriate knowledge distillation methods for NMT. Sentence-level distillation emerges as a robust choice for complex text scenarios, while token-level distillation is preferable in simpler, less complex environments.

\subsection{Impact of Decoding Difficulty}
\label{subsec:decoding_strategy}

In this subsection, we examine the relationship between decoding difficulty and the performance of knowledge distillation methods. For decoding methods, we mainly take teacher forcing~\cite{toomarian1992learning,lamb2016professor} and beam search~\cite{jaszkiewicz1999light} into consideration.
Beam search explores multiple hypotheses at each decoding step conditioned on the previous decoding results.
Teacher forcing, different with beam search, directly uses the previous target sequence as condition at each step of sequence generation, effectively preventing error amplification during decoding. This method simplifies the decoding process and can lead to improved performance~\cite{baskar2019promising}, which can be regarded as a simpler scenario in terms of decoding methods compared with the beam search.

\subsubsection{Experimental Setup and Methodology}
\label{subsec:decoding_difficulty_exp_setup_methodology}
Experiments are conducted using the same datasets and teacher models as in Tables \ref{tab:combined_impact_of_model_capacity} and \ref{tab:combined_impact_of_data_complexity_extended}.
The focus of our experiments is to closely examine the performance of token-level distillation and sentence-level distillation under different decoding difficulties (i.e., teacher forcing and beam search methods) on each dataset. Specifically, during the prediction phase, we employ beam search (BS) and teacher forcing (TF) methods. The former method considers the most probable candidates at each step of word generation, selecting one to include in the final sentence output. The latter method inputs the actual previous word into the model, rather than the model's own prediction from the previous step.

\subsubsection{Analysis of Results and Summary}
\label{subsec:decoding_difficulty_analysis_summary}
Table \ref{tab:combined_impact_of_Decoding_Difficulty} presents a comparison of BLEU scores for both BS and TF methods across token-level and sentence-level distillation. 
Our results indicate that teacher forcing is more effective at the token-level compared to the sentence-level, as evidenced by the negative values in $\bigtriangleup$TF across all datasets. This suggests that token-level distillation is better suited for the teacher forcing decoding approach. 

Conversely, in the more complex beam search scenario, sentence-level distillation tends to outperform token-level distillation, as indicated by the positive values in $\bigtriangleup$BS. This shift in effectiveness from token-level in TF to sentence-level in BS aligns with our hypothesis that teacher forcing, being a simpler decoding method, is more effective in scenarios where the decoding process is less complex. The token-level distillation benefits from the simplicity of the teacher forcing method, as it allows seeing the correct prefix words during decoding, making the process simpler and thus more effective.

\subsection{Summary}
Based on our three comprehensive analyses  focusing on model size, text complexity, and decoding difficulty, we have observed that token-level distillation is generally more suitable for scenarios involving larger student models, simpler texts, and greater amounts of available decoding information. In contrast, sentence-level distillation tends to be more effective in scenarios with smaller student models, more complex texts, and limited decoding information. These findings align with our initial hypothesis, suggesting that token-level distillation is better suited for simpler scenarios, while sentence-level distillation is more adept at handling complex situations.

\section{Hybrid Method for Combining Token-Level and Sentence-Level Distillation}
Despite our experimental results validate our hypothesis regarding the effectiveness of token-level and sentence-level distillation in different scenarios, we face the challenge of accurately defining the complexity level of each scenario. This issue complicates the optimal application of distillation methods in neural machine translation (NMT). In response, we propose a hybrid method, which combines token-level and sentence-level distillation through a dynamic gating mechanism. This method is designed to utilize the strengths of both distillation strategies and be adaptable across various scenarios, ranging from ``simple'' to ``complex''.

\subsection{Hybrid Distillation Method}
\label{subsec:integrated_methodology}
\begin{figure}
\centering 
\includegraphics[width=0.8\linewidth]{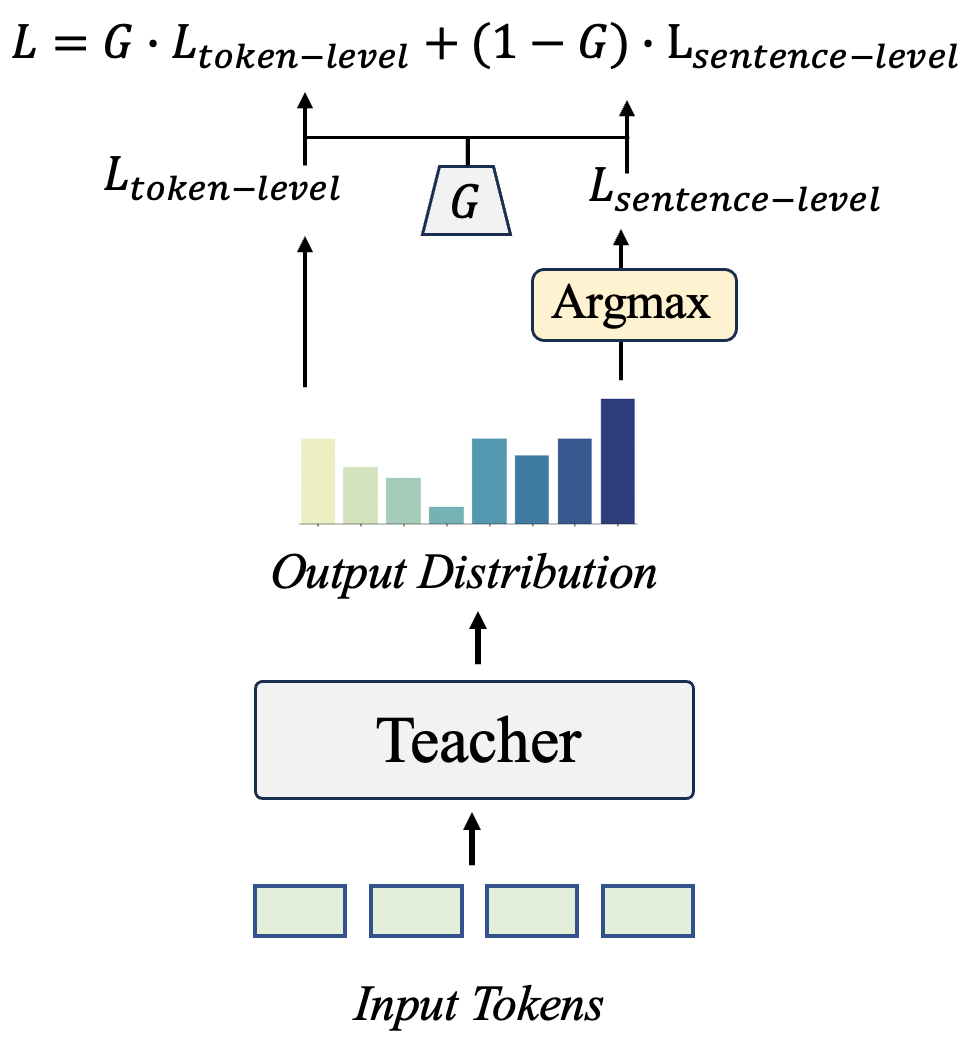} 
\caption{Architecture of the hybrid distillation method.}
\label{Fig.adaptive}
\end{figure}

Our hybrid method features a gate-controlled mechanism, dynamically balancing the contributions of token-level and sentence-level distillation. This mechanism, denoted as \( G \) and illustrated in Figure~\ref{Fig.adaptive}, is represented by the function \( g(x) \) for each input sequence \( x \), modulating the influence of each distillation strategy during training to suit different translation scenarios.

The overall loss function, \( L \), is a hybrid of token-level and sentence-level distillation losses, modulated by \( G \). Let \( x = \{x_1, \ldots, x_n\} \) and \( y = \{y_1, \ldots, y_m\} \) respectively represent the input and output (target) sequences. The probabilities \( \mathrm{P}_{\mathrm{s}}(y_j \mid x) \) and \( \mathrm{P}_{\mathrm{t}}(y_j \mid x) \) represent the output probabilities at position \( j \) for the student and teacher models, respectively.

For each input sequence \( x \), the gate-controlled parameter \( g(x) \) is defined as:
\begin{equation}
g(x) = \frac{1}{1 + e^{-z(x)}}
\end{equation}
where \( z(x) \) is a function of the input sequence \( x \), determining the balance between token-level and sentence-level distillation for that particular input.

The token-level loss \( L_{\text{token-level}}(x) \) is defined as:
\begin{equation}
L_{\text{token-level}}(x) = -\sum_{j=1}^m \sum_{y_j \in V} \mathrm{P}_{\mathrm{t}}\left(y_j \mid x\right) \log \mathrm{P}_{\mathrm{s}}\left(y_j \mid x\right)
\end{equation}
which sums over all tokens \( y_j \) in the vocabulary \( V \), weighted by the probability of teacher model \( \mathrm{P}_{\mathrm{t}}(y_j \mid x) \) and the logarithm of the probability of student model \( \mathrm{P}_{\mathrm{s}}(y_j \mid x) \).

The sentence-level loss \( L_{\text{sentence-level}}(x) \) is defined as:
\begin{equation}
L_{\text{sentence-level}}(x) = -\log \mathrm{P}_{\mathrm{s}}(\hat{y} \mid x)
\end{equation}
which is the negative logarithm of the student model's probability of the actual output sequence \( \hat{y} \) given by the teacher model.

Therefore, the overall loss function \( L \) for an input sequence \( x \) is given by:
\begin{equation}
L(x) = g(x) \cdot L_{\text{token-level}}(x) + (1 - g(x)) \cdot L_{\text{sentence-level}}(x)
\end{equation}
This formulation allows \( L(x) \) to represent the combined loss for a given input sequence \( x \), effectively integrating the token-level and sentence-level distillation losses. By dynamically adjusting the weights of token-level and sentence-level distillation through \( g(x) \), our hybrid method adapts to different input sequences, enhancing the effectiveness of model training.

\subsection{Implementation Details}
\label{subsubsec:implementation_details}
The training process begins with training a BiBERT teacher model at its base size to generate reference outputs. Subsequently, we implement our hybrid distillation method. This approach allows the model to adaptively switch between token-level and sentence-level strategies, optimizing the most effective learning path throughout the training process.
Our experiments are conducted on four NVIDIA 3090 GPUs, each with a batch size of 3000. Gradients accumulate over four iterations per update. The learning rate is set at \(5 \times 10^{-4}\), using the Adam optimizer with an inverse-sqrt learning rate scheduler. For inference, we employ a beam search with a width of 4 and a length penalty of 0.6.

\subsection{Baselines}
\label{subsec:baseline_methods}
In our study, we compared our hybrid distillation approach with several advanced baseline methods in NMT:

\begin{itemize}
  \item \textbf{Transformer + R-Drop}~\cite{wu2021r}: Utilizes regularization to minimize the bidirectional KL-divergence between sub-models' outputs.
  \item \textbf{CipherDAug}~\cite{kambhatla2022cipherdaug}: Employs a novel data augmentation technique based on ROT-k ciphers.
  \item \textbf{Cutoff}~\cite{shen2020simple}: Implements a data augmentation strategy that erases part of the information within an input sentence.
  \item \textbf{Cutoff+Knee}~\cite{iyer2023wide}: Combines Cutoff with an Explore-Exploit learning rate schedule.
  \item \textbf{SimCut and Bi-SimCut}~\cite{gao2022bi}: Enforces consistency between the output distributions of original and cutoff sentence pairs.
  \item \textbf{Transformer + R-Drop + Cutoff}~\cite{wu2021r}: Integrates R-Drop regularization with Cutoff data augmentation.
  \item \textbf{Cutoff + Relaxed Attention + LM}~\cite{lohrenz2023relaxed}: Introduces relaxed attention as a regularization technique.
  \item \textbf{BiBERT}~\cite{DBLP:conf/emnlp/XuDM21}: Utilizes a bilingual pre-trained language model for the NMT encoder.
\end{itemize}

\subsection{Experimental Results}
\label{subsec:main_results}

\begin{table}[htbp]
  \centering
  \caption{Experimental results on IWSLT14 de$\rightarrow$en of baseline methods and our hybrid method.}
  \resizebox{\columnwidth}{!}{
  \begin{tabular}{lc}
    \toprule
    Methods & BLEU \\
    \midrule
    Transformer + R-Drop~\cite{wu2021r} & 37.25 \\
    CipherDAug~\cite{kambhatla2022cipherdaug} & 37.53 \\
    Cutoff~\cite{shen2020simple} & 37.60 \\
    Cutoff+Knee~\cite{iyer2023wide} & 37.78 \\
    SimCut~\cite{gao2022bi} & 37.81 \\
    Transformer + R-Drop + Cutoff~\cite{wu2021r} & 37.90 \\
    Cutoff + Relaxed Attention + LM~\cite{lohrenz2023relaxed} & 37.96 \\
    Bi-SimCut~\cite{gao2022bi} & 38.37 \\
    BiBERT~\cite{DBLP:conf/emnlp/XuDM21} & 38.61 \\
    \midrule
    Our Hybrid Distillation & \textbf{39.30} \\
    \bottomrule
  \end{tabular}
    }
  \label{tab:main_experiment_results}
\end{table}
Table~\ref{tab:main_experiment_results} shows the translation accuracy (indicated by BLEU score) of our method and baseline methods. The results demonstrate that our hybrid distillation method outperforms all baseline models, achieving a BLEU score of 39.30, which indicates the efficiency of our method in combining token-level and sentence-level distillation strategies.

\subsection{Ablation Study}
\label{subsec:ablation_study}
The ablation study evaluates the individual impacts of token-level and sentence-level distillation within our hybrid method, aiming to understand their contributions to the overall translation performance.

\begin{table}[htbp]
  \centering
  \caption{Ablation study results of distillation methods on IWSLT14 de$\rightarrow$en.}
  \begin{tabular}{lcc}
    \toprule
    Methods & Model Params & BLEU \\
    \midrule
    Sentence-Level & 78M & 39.01 \\
    Token-level & 78M & 39.15 \\
    Our Hybrid Distillation & 78M & 39.30 \\
    \bottomrule
  \end{tabular}
  \label{tab:ablation_results}
\end{table}

Table~\ref{tab:ablation_results} presents the results of the ablation study. The individual performances of sentence-level and token-level distillation highlight their respective strengths in enhancing translation quality. The sentence-level method, with a BLEU score of 39.01, demonstrates its capability in capturing the overall semantic coherence, while the token-level method, scoring slightly higher at 39.15, shows its effectiveness in ensuring precise token-level translations. Our hybrid method, achieving a BLEU score of 39.30, surpasses these individual strategies, indicating that the synergistic combination of token-level precision and sentence-level coherence can yield superior results. The results show our hybrid method, which combines token-level and sentence-level distillation, effectively navigates the challenges in scenarios with ambiguous complexity levels, enhancing translation quality in neural machine translation.

\subsection{Analysis of Gate-Controlled Mechanism}
\begin{figure}[t]
\centering 
\includegraphics[width=\linewidth]{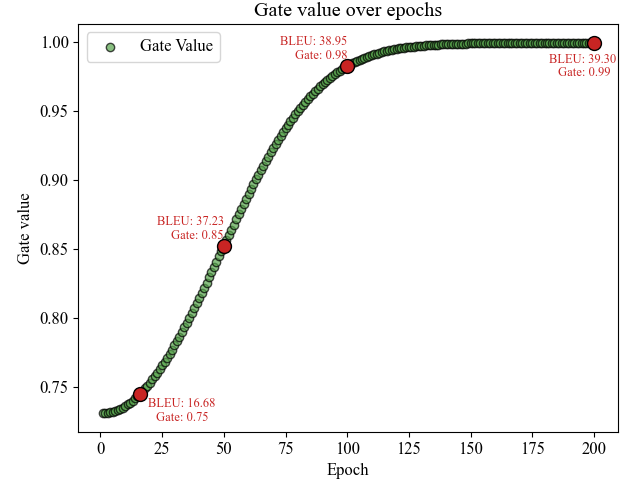} 
\caption{Dynamics of gate value \( G \) over training epochs.}
\label{Fig.gata}
\end{figure}
To understand the learning process of the learnable gate-controlled mechanism \( G \) and to verify the effectiveness of this learning method, we present the dynamics of the gate value \( G \) over training epochs during our experiments, as shown in Figure \ref{Fig.gata}. We find that at the beginning of the learning process of \( G \), its value is around 0.72. As training progresses (around 20 epochs), the value of \( G \) increases to 0.75, with the corresponding BLEU score being 16.68. With further training (around 50 epochs), \( G \) gradually rises to 0.85, and the BLEU score significantly improves to 37.23. During this phase, the increase in the value of \( G \) is quite apparent, and there is a notable enhancement in the BLEU score. Subsequently (around 100 epochs), \( G \) increases to 0.98, and the BLEU score rises to 38.95. At this stage, although \( G \) continues to increase, the growth rate of the BLEU score slows down compared to the previous phase. Eventually, the value of \( G \) approaches 1, and the BLEU score reaches 39.30. We believe that initially, sentence-level learning is easier, while token-level learning is more challenging. Therefore, the model first learns the simpler aspects, leading to a faster increase in the BLEU score. As the simpler tasks are mastered, the model then moves on to the more difficult token-level learning, resulting in a slower rate of improvement in the BLEU score. 
From the results, it is evident that the learnable parameters, by adjusting the size of \( G \), effectively enable the model to autonomously learn knowledge from sentence-level distillation and token-level distillation, demonstrating the effectiveness of our design.

\section{Conclusion}
In this paper, we conduct an in-depth exploration of the two main methods of knowledge distillation in neural machine translation (NMT): sentence-level and token-level distillation. We hypothesize that token-level distillation is more suitable for simpler scenarios, whereas sentence-level distillation is better for complex scenarios. To test this hypothesis, we systematically analyze the impact of varying the size of the student model, the complexity of the text, and the difficulty of the decoding process. Our empirical results validate our hypothesis, showing that token-level distillation generally performs better in scenarios with larger student models, simpler texts, and higher availability of decoding information (making decoding easier). In contrast, sentence-level distillation performs better in scenarios with smaller student models, more complex texts, and limited decoding information (making decoding harder). To address the challenge of defining the difficulty level of specific scenarios, we further introduce a dynamic gate-controlled mechanism that combines the advantages of both token-level and sentence-level distillation. Our experiments validate the effectiveness of this hybrid method over the single distillation method and baselines methods.

\section*{Ethical Statement}

There are no ethical issues.

\section*{Acknowledgments}
We are grateful to the anonymous reviewers of IJCAI for their constructive comments that significantly improve the manuscript. This work is supported by the Liaoning Provincial Applied Basic Research Program, grant number 2022JH2/101300258.

\bibliographystyle{named}
\bibliography{ijcai24}

\end{document}